\documentclass[sigconf]{acmart}

\usepackage{booktabs} 
\usepackage{hyperref}

\setcopyright{rightsretained}

\acmDOI{}

\acmISBN{}

\acmConference[Neuromorphic Computing]{ORNL Neuromorphic workshop}{July 2017}{Oak Ridge, Tennessee USA} 
\acmYear{2017}
\copyrightyear{2017}

\acmPrice{}

\begin{document}
\title[Community detection with spiking neurons]{Community detection with spiking neural networks for neuromorphic hardware}
\titlenote{This work was supported by the United States Department of Defense and used resources of the Computational Research and Development Programs at Oak Ridge National Laboratory. This manuscript has been authored by UT-Battelle, LLC, under Contract No. DE-AC0500OR22725 with the U.S. Department of Energy. The United States Government retains and the publisher, by accepting the article for publication, acknowledges that the United States Government retains a non-exclusive, paid-up, irrevocable, world-wide license to publish or reproduce the published form of this manuscript, or allow others to do so, for the United States Government purposes. The Department of Energy will provide public access to these results of federally sponsored research in accordance with the DOE Public Access Plan.}

\author{Kathleen E. Hamilton}
\orcid{0000-0001-6382-5665}
\affiliation{%
  \institution{Oak Ridge National Laboratory \\ Computing \& Computational Sciences Dir}
  \streetaddress{One Bethel Valley Road}
  \city{Oak Ridge} 
  \state{Tennessee} 
  \postcode{37831-6015}
}
\email{hamiltonke@ornl.gov}

\author{Neena Imam}
\affiliation{%
  \institution{Oak Ridge National Laboratory \\ Computing \& Computational Sciences Dir}
  \streetaddress{One Bethel Valley Road}
  \city{Oak Ridge} 
  \state{Tennessee} 
  \postcode{37831-6015}
}
\email{imamn@ornl.gov}

\author{Travis S. Humble}
\affiliation{%
  \institution{Oak Ridge National Laboratory \\ Computing \& Computational Sciences Dir}
  \streetaddress{One Bethel Valley Road}
  \city{Oak Ridge} 
  \state{Tennessee} 
  \postcode{37831-6015}
  }
\email{humblets@ornl.gov}

\renewcommand{\shortauthors}{K. Hamilton et al.}

\begin{abstract}
We present results related to the performance of an algorithm for community detection which incorporates event-driven computation. We define a mapping which takes a graph $\mathcal{G}$ to a system of spiking neurons. Using a fully connected spiking neuron system, with both inhibitory and excitatory synaptic connections, the firing patterns of neurons within the same community can be distinguished from firing patterns of neurons in different communities. On a random graph with $128$ vertices and known community structure we show that by using binary decoding and a Hamming-distance based metric, individual communities can be identified from spike train similarities.  Using bipolar decoding and finite rate thresholding, we verify that inhibitory connections prevent the spread of spiking patterns. 
\end{abstract}

%
%



\maketitle

\section{Introduction}
\label{sec:Introduction}
Graph partitioning and community detection are ubiquitous tasks encountered in a diverse set of sciences and many methods have been developed to sort the vertices of a graph, or nodes of a network into classes based on similarity measures. These methods utilize graphical characteristics and structures, such as transitivity, modularity or betweenness, and spectral-based analysis and often require large scale matrix analysis \cite{boccaletti2006complex,fortunato2010community,malliaros2013clustering,PhysRevE.70.066111}. These methods can be parallelized and many algorithms exist for the analysis of very large networks. However, the emergence of unconventional processors, such as neuromorphic processors, requires approaches which utilize event-based computation. We present work in this paper related to the development of an algorithm which incorporates event-based computation in the identification of related vertices in networks or graphs.  

Inspired by the recent work of Shaub et al \cite{Schaub2017} in which the different approaches to community detection are organized according to the problem they are designed to solve (e. g. partitioning problems, clustering problems, dynamical problems), we describe our approach as a hybrid dynamical clustering method which incorporates the discrete time signals of a spiking neuron system. In this paper we present results that serve as a proof-of-concept related to the mapping of recurrent neural networks based on interacting spin dynamics (Hopfield networks) to spiking neural systems.

Our goal is to construct a system of spiking neurons which can be used to generate a set of spike responses which can identify vertex communities in a graph. We choose to characterize a community in a graph $\mathcal{G}(\mathbb{V},\mathbb{E})$ as a subset of vertices $v \in \mathbb{V}$ such that the density of edges internal to this subset is higher than the density of edges connecting to the remainder of the graph. This is similar to definitions used in other cluster based models \cite{NewmanNetworks}. There is a well known aphorism in Hebbian learning, attributed to Siegrid L\"{o}wel \cite{Lowel209}: ``neurons that fire together are wired together.'' We use the statement inversion:``neurons that are wired together, fire together'' to construct our approach to community detection. 

The vertices and edges of a given graph $\mathcal{G}$ are mapped to a network of symmetrically connected spiking neurons, which is then selectively driven by time-dependent external currents. There must be a degree of similarity between the resulting spike trains which can be used to distinguish individual communities of $\mathcal{G}$. We binary decoding and bipolar decoding of spike trains with similarity between trains measured using a Hamming distance metric. The ability to identify individual communities is dependent on the linear separability of the Hamming distance metric values. This is controlled by the size of the bin width $\Delta t$ used in the binary decoding of individual spike trains. 

This approach incorporates Hopfield networks \cite{hopfield1982neural,hopfield1984neurons,hopfield1985neural} and spin glass models \cite{reichardt2004detecting,PhysRevE.74.016110,PhysRevLett.76.3251}.  It has been shown that recurrent networks with steady states can be used to find solutions to the problem of graph partitioning \cite{hertz1991introduction,van1990graph}. The use of positive and negatively weighted edges is needed to drive the system to a solution which meets the optimization conditions. However, these approaches are often limited to finding partitions of equal sizes in a graph, or require prior knowledge of the number of communities to find. 

Hopfield networks have been implemented using spiking neurons for the task of content addressable memories and pattern retrieval \cite{tanaka2005associative,maass1997networks}, in this work we focus on the application of the recurrent neural network model to a task related to graph partitioning. We show in this paper that in conjunction with carefully chosen parameters for our neuron model, these spin-glass based models can be used to find un-equal sized groups. We establish that spiking data can be used to identify  communities.

Recent works on community detection have focused on how real-world networks may have ambiguous community structure and there are limitations how much information can be inferred from metadata based approaches \cite{peel2017ground}. In this work we focus on a discussion of how spiking data can be used to identify communities and we work with graphs generated with known communities and fixed labels. The graphs analyzed in this paper are all instances of a specific class of random graphs: Girvan-Newman benchmark graphs \cite{girvan2002community}. These graphs have $128$ vertices organized into $4$ equal-order communities of $32$ vertices each. The average degree is $\langle d \rangle = 16$. Using these graphs we demonstrate our spike-based approach to community detection, focusing on binary decoding and bipolar decoding of spike trains. 

A graph is mapped to a system of spiking neurons and driven in such a manner that the generated spike trains can be used to reconstruct the known community labels of the original graph. Two neurons ($n_i, n_j$) in $Q_{i=j}$ must have firing patterns that exhibit a degree of similarity which distinguishes them from the spike trains generated by neurons ($n_i, n_j$) in $Q_{j\neq i}$. In Section \ref{sec:SNN_construction}, we derive the theoretical structure which underlies the main components of our approach. We begin with how a graph $\mathcal{G}$ is mapped to a spiking neural network (SNN). Then, we discuss how the physical parameters associated with the spiking neuron system must be set and how the selective driving of this SNN must be done in order to generate a set of spike trains are generated that can distinguish a single community $Q_i$ from the remaining $Q_{j \neq i}$.  In Sections \ref{sec:binary_code} and \ref{sec:bipolar_code} we describe how spike responses are decoded and we discuss the metrics we use to analyze decoded spike trains. This paper presents on results which establish proof-of-concept that a mapping and driving pattern exists which can be used to generate spike trains characteristic of a graph's known community structure. In Section \ref{sec:future_work} we introduce how our spiking neural networks can be incorporated into existing community detection algorithms.

\section{Spiking neuron model construction}
\label{sec:SNN_construction}
We focus on community detection in undirected, unweighted graphs. A graph is defined by a vertex set and an edge set $\mathcal{G} = \mathcal{G}(\mathbb{V},\mathbb{E})$. Multiple edges and self-loops are not allowed. In this initial work, we study graphs with clearly delineated (non-overlapping) communities. 

SNNs are dynamical systems which compute without the use of steady states \cite{maass1997networks,maass2001pulsed}. Information is transmitted through electrical pulses. The neurons which compose the spiking network are nonlinear units and can exhibit a rich set of dynamics and firing patterns based on the physical parameters. We use few parameters to build our SNNs; they are leaky-integrate and fire neurons defined by a \textit{threshold voltage} ($v_{th}$), a \textit{refractory period} ($t_{r}$), and a \textit{time constant} ($\tau$). The full spiking network itself $S(\mathbb{N},\mathbb{W})$ is defined by a set of homogeneous neurons $\mathbb{N} = \lbrace n_i \rbrace$, and a set of symmetrically connected synapses$\mathbb{W} = \lbrace s(w)_{ij} \rbrace$, weighted edges which connect neurons $n_i \leftrightarrow n_j$. 

For a graph $\mathcal{G}(\mathbb{V},\mathbb{E})$, we assume there is a set of known vertex communities $\lbrace Q_i \rbrace$ such that $\bigcup_i Q_i = \mathbb{V}(\mathcal{G})$ and no vertex exists in more than one community.  The remainder of this paper will develop how spiking systems can identify individual communities $Q_i$.

The mapping of an undirected, unweighted graph to SNN is analogous to the construction of a Hopfield recurrent neural network; using symmetric connections which are either positive or negatively weighted. The two-step mapping first defines a SNN by defining a spiking neuron for each vertex on the graph: $v_i \in \mathbb{V} \to n_i \in \mathbb{N}$, and each edge on the graph defines a symmetric pair of excitatory synapses $e_{ij} \in \mathbb{E} \to s(w_{+})_{ij},s(w_{+})_{ji} \in \mathbb{W}, w_{+}>0$. This SNN is then transformed to a fully connected SNN by the addition of symmetric pairs of inhibitory synapses $s(w_{-})_{jk},s(w_{-})_{kj} \in \mathbb{W}, w_{-}<0$ for any edges which do not exist on the original graph. The magnitudes of the excitatory and inhibitory synapses are equal $|s(w_{+})| = |s(w_{-})|$.

Leaky integrate and fire neurons are simplified models of neuronal behavior \cite{gerstner2002spiking}. We work with these models because of their close proximity to the behavior of the IBM TrueNorth processor \cite{merolla2014million,cassidy2013cognitive}. The resulting networks of spiking neurons do not require extensive detail about the biological behavior, nor are they an attempt to describe cortical network dynamics. We focus on a model of the membrane potential in which the potential $v_i(t)$ of a single neuron is a time-dependent function which changes depending on discrete or continuous inputs. Discrete signals are measured upon the arrival of positively and negatively weighted spikes $\Delta v = s(w_{\pm})$, any external driving force term in $V_{ext}(t)$ is assumed to be continuous. Additionally, we use the existence of a non-zero leak $\tau$ to continuously relax $v(t)$ to an equilibrium value $v_r$.
\begin{align}
\frac{d(v_i(t))}{dt} &= \frac{V_{ext}(t) - v_i(t)}{\tau}  \\
V_{ext}(t) &= I_{ext}(t)R + \sum_{t^{\prime},j\to i} s(w)_{ij} \delta(t-t^{\prime})
\label{eq:LIF_EOM}
\end{align}

If a neuron's potential exceeds the firing threshold $v_{th}$, it fires a spike along its synaptic connections and enters a refractory period $t_{R}$ during which its potential is not changed. The choice of the neuron system parameters is determined by whether they are used to generate similar spike trains, or to enhance dissimilarity between spike trains. The refractory period $t_R$ is needed to impose a sense of directionality to the spread of spiking patterns. Aided by the positively weighted synapses, spike responses spread to neurons that have not yet fired.  These correspond to connected vertices on the graph and lead to similarities in spike trains of neurons $n_i, n_j$ in the same community $Q_i$. The parameters that lead to dissimilar spike trains are those which help to impede the spread of spiking patterns across several neurons in a system, primarily the negatively weighted synaptic connections. Many parameters play a dual role, necessary for both creating similar spike trains and to inhibit spike pattern spread. The time constant ($\tau$) is balanced to ensure arriving spikes can accumulate and lead to secondary firing while also ensuring that the effects of spike impulses are not long-lived, the firing threshold $v_{th}$ ensures that a neuron fires in response to incoming spike impulses only when multiple impulses arrive in a short time window. 

For a set of homogeneous spiking neurons, there are many system parameters which are tunable in order to generate spiking patterns which are informative about the original graph's community structure. Previously, we have studied how the simplest Hopfield network could be mapped to a system of spiking networks \cite{Hamilton2017ns17} we considered graphs with fully connected communities connected by a single bridge bond (barbell graphs). These networks could be driven using a pair of sinusoidal currents out of phase by 180 degrees. The negative driving current, as well as a careful tuning of neuron system parameters (refractory period, time constant, firing threshold and synaptic weight) was sufficient to generate spike trains characteristic of the two communities. In this work we generalize our approach to a system connected as a spin glass, the role of the negative driving current is replaced by the negatively weighted synapses, which inhibit the growth and limit the occurrences of spike cascades.

The neuron dynamics under square pulse driving are relatively simple. Only one neuron is actively driven by a square pulse at any time ($t$), and the effects from the refractory period $t_R$ allow for the equations of motion in Eq. \eqref{eq:LIF_EOM} to be separated as: a set of equation to describe the firing dynamics under active driving, and one set of equations to describe the firing dynamics as a reaction to a neighbor being actively driven. The equation for the square pulse is,
\begin{equation}
V_{ext}(t) = A_{max} \left[ \tanh{(\beta [t - t_1])} - \tanh{(\beta[t_2 - t])} \right]
\end{equation}
The shape of the square pulse is determined by the pulse height ($A_{max}$), pulse width ($t_A = t_2 - t_1$), and the parameter $\beta$ which determines the sharpness of the pulse's rise. The gap between subsequent pulses applied to different neurons is sufficiently large such that any $\Delta V$ induced by $V_{ext}$ on neuron $n_i$ has decayed away before neuron $n_j$ is actively driven. The parameter $\beta$ controls the sharpness of the square pulse. A sharp step is needed to ensure that spikes are fired only when the external current is a constant driving force, $I_{max}= 2 A_{max}/R$, however the function $V_{ext}(t)$ must remain continuous. Under active square pulse driving, effect of the constant driving force leads to a constant firing rate ($\delta$), which can be found by integrating the equation of motion:
\begin{equation}
\frac{d(v(t))}{dt} = \frac{V_{ext}(t) - v(t)}{\tau},
\end{equation}
and is found in terms of the time constant $\tau$, the square pulse amplitude $A_{max}$, the reset voltage $v_0$ and the spike threshold voltage $v_{th}$:
\begin{equation}
\frac{\delta}{ \tau} = \log{\left ( \frac{2 A_{max} - v_0}{2 A_{max} - v_{th} }\right)}.
\end{equation}
The firing rate $\delta$ can be used to determine the synaptic weight $s_w$ with this assumption that a neuron ($n_j$) will fire a spike in response to one of its nearest neighbors ($n_i$, $s(w_{+})_{(i \to j)}>0$) being actively driven when $2$ spikes arrive at $n_j$ in a short time span.  Setting $s_w = \alpha v_{th} (0 < \alpha < 1)$, the first spike arrives at $t_0$ and increases the potential $v_j(t_0) = v_0 + s(w_{+})$. The arrival of the second spike happens at $t = \delta$ and the resulting potential must exceed the spike threshold: 
\begin{equation}
v_j(t_0+\delta) > v_{th}. 
\label{eq:multispike_cond}
\end{equation}Again integrating the dynamical equation for $v_j$  during time $t : [t_0,t_0+\delta]$ is straightforward since there is no active driving. The inequality in Eq. \eqref{eq:multispike_cond} becomes:
\begin{align}
& s_w\left(e^{-\delta / \tau} + 1\right) > v_{th}, \\
& \frac{\delta}{\tau} < \log{ \left(\frac{\alpha}{1+\alpha}\right)}.
\end{align}

\section{Binary decoding and spike train similarity} 
\label{sec:binary_code}
For each neuron $n_i \in \mathbb{N}(\mathcal{S})$, there is an associated set of spike times called a spike train. The spike train data is analyzed using a comparison matrix, where similarity between trains is quantified by a similar metric as one used in Ref. \cite{humphries2011spike}. Binary decoding converts the spike trains to binary vectors $x_i^{(m)}(\Delta t)$ using a discrete time step $\Delta t$; $x_i$ has a value of $1$ if at least one spike occurs in the time window $\Delta t$. The label $(m)$ assigned to a given neuron is included for completeness. The set of all $\lbrace x_i^{(m)}(\Delta t) \rbrace$ binary vectors of length $|x_i|$ are compared pairwise to construct entries in an $|\mathbb{N}| \times |\mathbb{N}|$ comparison matrix $H$ with entries $H_{ij}$ defined by the normalized Hamming distance between two binary decoded spike trains: $H_{ij} = 1-h(x_i^{(n)},x_j^{(m)})/|x_i|$ \cite{humphries2011spike}. We introduce a modified version of this comparison matrix:
\begin{equation}
H_{ij} = \left ( 1 - \frac{h\left [x_i^{(n)}(\Delta t),x_j^{(m)}(\Delta t)\right]}{|x_i|} \right)(\mathbf{1}^{T} \mathbf{x}_i)(\mathbf{1}^{T} \mathbf{x}_j).
\label{eq:Hamming_metric_modified}
\end{equation}
The inclusion of the terms: $\mathbf{1}^{T} \mathbf{x}_i,\mathbf{1}^{T} \mathbf{x}_j$ down weights the entries of the matrix $H_{ij}$ in which the spike train of a firing neuron is compared to a non-firing neuron (e.g. $x_i = \lbrace 0 \rbrace^{\otimes |x_i|}$). Later (see Sec. \ref{sec:future_work}), as we begin to develop scalable methods for real-world analysis, these terms will be omitted and the Hamming metric will be defined as $0$ when either $x_i$ or $x_j$ corresponds to a non-firing neuron.

\begin{figure}
\includegraphics[width=0.9\columnwidth]{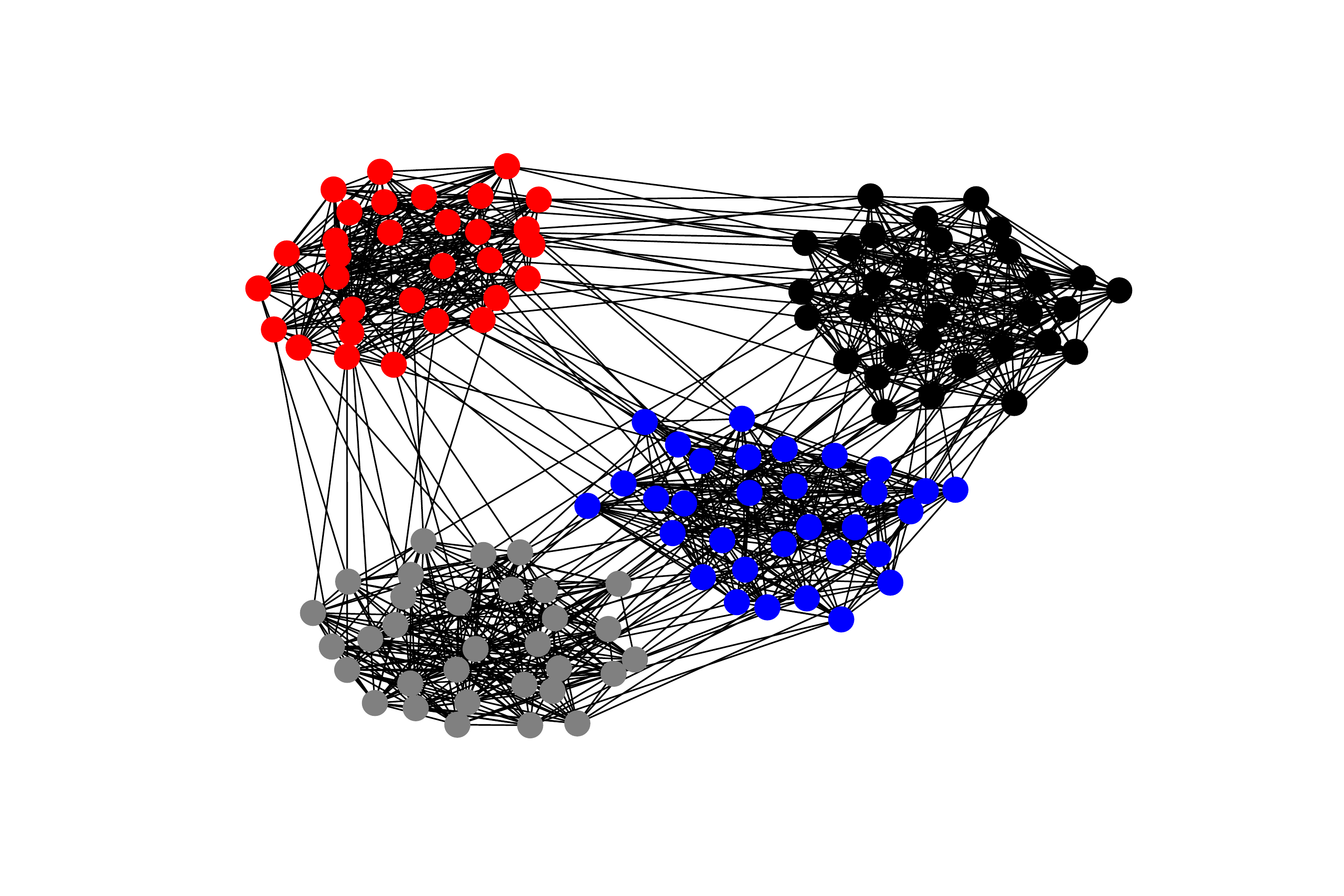}
\caption{The Girvan-Newman benchmark graph instance generated by software available from \cite{fortunato_site} and used to generate the spiking data analyzed in this paper.The different communities are distinguished by vertex color: Community 1 (red), Community 2 (blue), Community 3 (black), Community 4 (grey).}
\label{fig:girvan_newman_graph}
\end{figure}

In the initial tests of our algorithm, we use the benchmark graphs of Girvan-Newman and Fortunato (see \cite{girvan2002community,newman2004finding,lancichinetti2008benchmark,lancichinetti2009benchmarks}) which have equal-sized, strongly-connected communities with minimal overlap (see Fig. \ref{fig:girvan_newman_graph}). Using the software available at \cite{fortunato_site}, we generate instances of the random Girvan-Newman graphs (and the known community memberships), map them to a SNN of homogeneous neurons, and simulate the spiking dynamics using the Brian2 Python library \cite{goodman2008brian}.  A SNN is constructed with the parameters: $\tau = 25 \; \mathrm{ms}$, $v_{th} = 0.8 \mathrm{V}$, $|s(w_{+})|=|s(w_{-})|=0.75 \;\mathrm{V}$, $v_0 = 0 \; \mathrm{V}$ and $t_{R} = 20 \; \mathrm{ms}$. 

The spiking data analyzed in this paper are all generated from the same graph instance (shown in Fig. \ref{fig:girvan_newman_graph}). Three of the four communities are driven: $Q_1, Q_2, Q_4$.  Neurons are driven by individual square pulses of maximum height $A_{max} \; \mathrm{V}$, and width $ \tau_{A}= 200 \; \mathrm{ms}$ with a gap between subsequent pulses of $\Delta t_{pulse} = 800 \; \mathrm{ms}$. The complete set of neurons which are driven is ordered by community: $Q_1$ is driven first, then $Q_2$, then $Q_3$. During driving, the primary firing neuron fires 10 spikes, separated by nearly uniform time intervals: $\delta_1 \approx 21 \; \mathrm{ms}$. The secondary firing neuron fires 5 spikes at nearly uniform time intervals: $\delta_2 \approx 42 \; \mathrm{ms}$. Slight variations in the time interval between spikes are possible due to spikes fired when $V_{ext} \lessapprox I_{max}R$, and approximations introduced during numerical integration of the equations of motion.

\begin{figure}
\includegraphics[width=1.0\columnwidth]{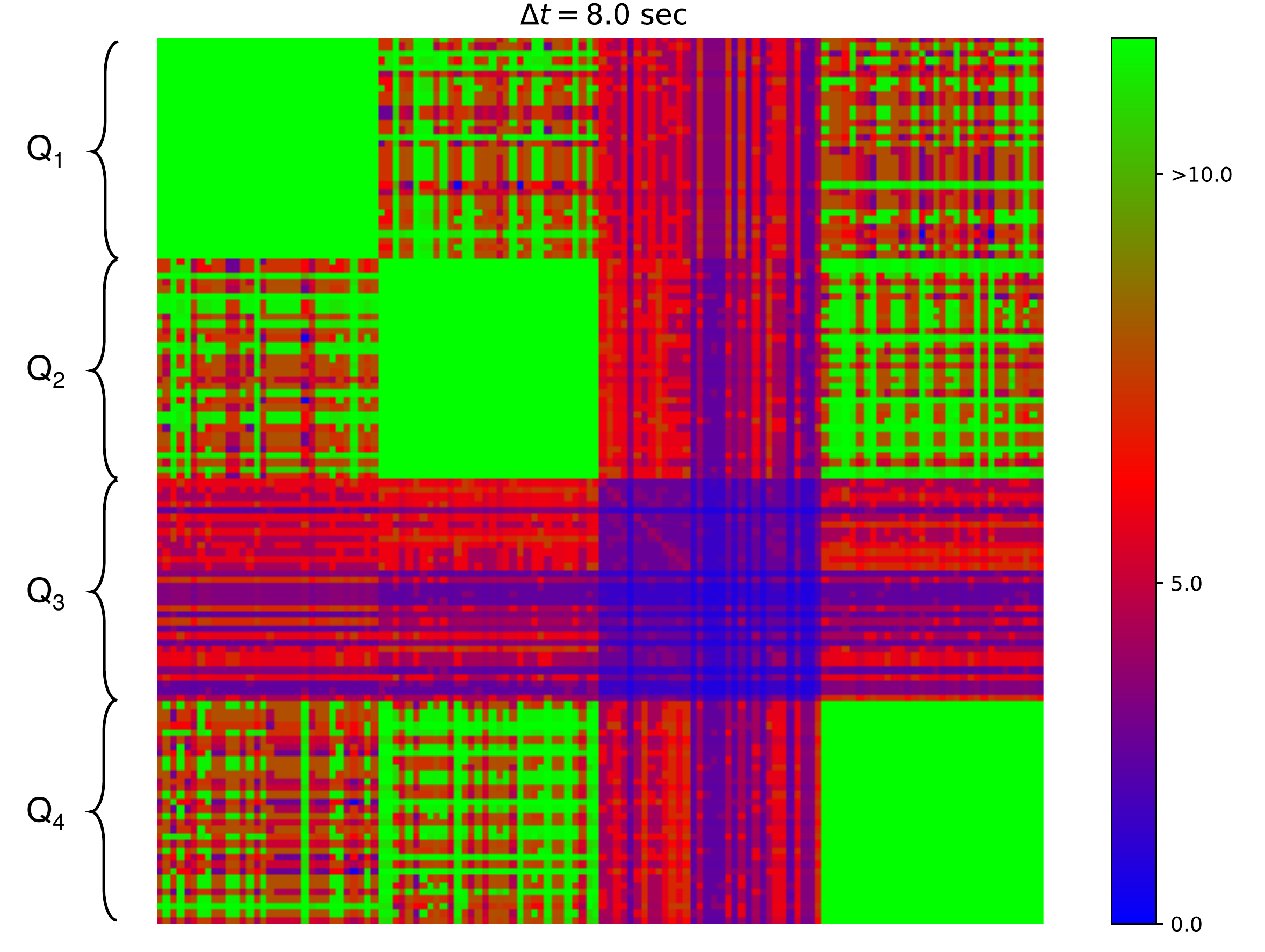}
\caption{The comparison matrix for a Girvan-Newman benchmark graph instance. Community three ($Q_3$)was not driven and is difficult to discern from the noisy background as a coherent spiking pattern is not able to spread from one community to the next. The Hamming metric used to generate this matrix included terms to down-weight low spiking neurons.}
\label{fig:comparison_matrix}
\end{figure}

From the generated spike trains, the comparison matrix is constructed using binary decoding with $\Delta t = 8.00 \; \mathrm{sec}$. In Fig. \ref{fig:comparison_matrix} the three driven communities can be identified by the magnitude of $H_{ij}$ and the block matrix structure. Along the diagonal, the value of $H_{ij}$ is significantly higher between a driven neuron $x_i^{(n)}$ and target neuron $x_j^{(n)}$ in the same community, compared to an off-diagonal block in which a driven neuron $x_i^{(n)}$ and target neuron $x_j^{(m)}$ are different communities. The un-driven neurons of $Q_3$ do not exhibit a strong similarity to any of the driven neurons in $Q_1, Q_2, Q_4$.

\section{Bipolar decoding and firing rate thresholding} 
\label{sec:bipolar_code}
The concept of a community in Section \ref{sec:Introduction} is now used to decode spiking output. During the community-ordered driving sequence used in Section \ref{sec:binary_code}, we conjecture that neurons undergoing active driving and contained in the same community will exhibit higher firing rates that those outside of the community. Only two neuronal states, corresponding to the Hopfield network states $+1$ and $-1$, are allowed. These states are mapped to spike train patterns through the use of time window binning. This is a variation on the binary code discussed in Section \ref{sec:binary_code} but now we consider the total number of spikes fired by a neuron during $\Delta t$. The Hopfield network state $+1$ is mapped to a neuron which has a firing rate over a fixed time window $\Delta t$ which exceeds a pre-determined threshold value $f_i \geq f_0$, this is considered to be an active state. The Hopfield network state $-1$ is mapped to a neuron which has a firing threshold over a fixed time window $\Delta t$ which does not exceed a pre-determined threshold value $f_i < f_0$, this is an inactive state. Similar approaches to mapping Hopfield networks to spiking neural networks have been investigated for the task of pattern retrieval using FitzHugh-Nagumo neuron models \cite{kanamaru2001fluctuation,kanamaru2000associative}. 

An upper bound on $f_0$ can be found by looking at the spiking response of a fully-connected $K_{32}$ community. For a fully connected clique, where each neuron is actively driven once, the maximum number of spikes that a neuron can fire is,
\begin{equation}
f_{max} = r_1 + (n-1)r_2,
\label{eq:max_count}
\end{equation}
where $r_1$ is the active firing rate and $r_2$ is the response firing rate. The minimum firing threshold value is still under investigation, but we use a simple heuristic to approximate $f_0$. For the Girvan-Newman benchmark graphs studied in this paper, the communities are known to each be of order $32$ and $\langle d \rangle =16$. For a lower bound on the minimum number of spikes a neuron can fire and still be counted as being a member of a community, we make the assumption that at least half of a vertex's neighbors must be in the same community. We replace the factor of $(n-1)$ in Equation \eqref{eq:max_count} with $16/2 + 1$,
\begin{equation}
f_{min} = r_1 + 9 r_2.
\end{equation}
Further study into a robust lower bound would require a stricter definition of what is a community. For our task of reconstructing a known set of labels, and with the non-overlapping community structure of the Girvan-Newman benchmarks, this heuristic will suffice.
\begin{figure}
\includegraphics[width=1.0\columnwidth]{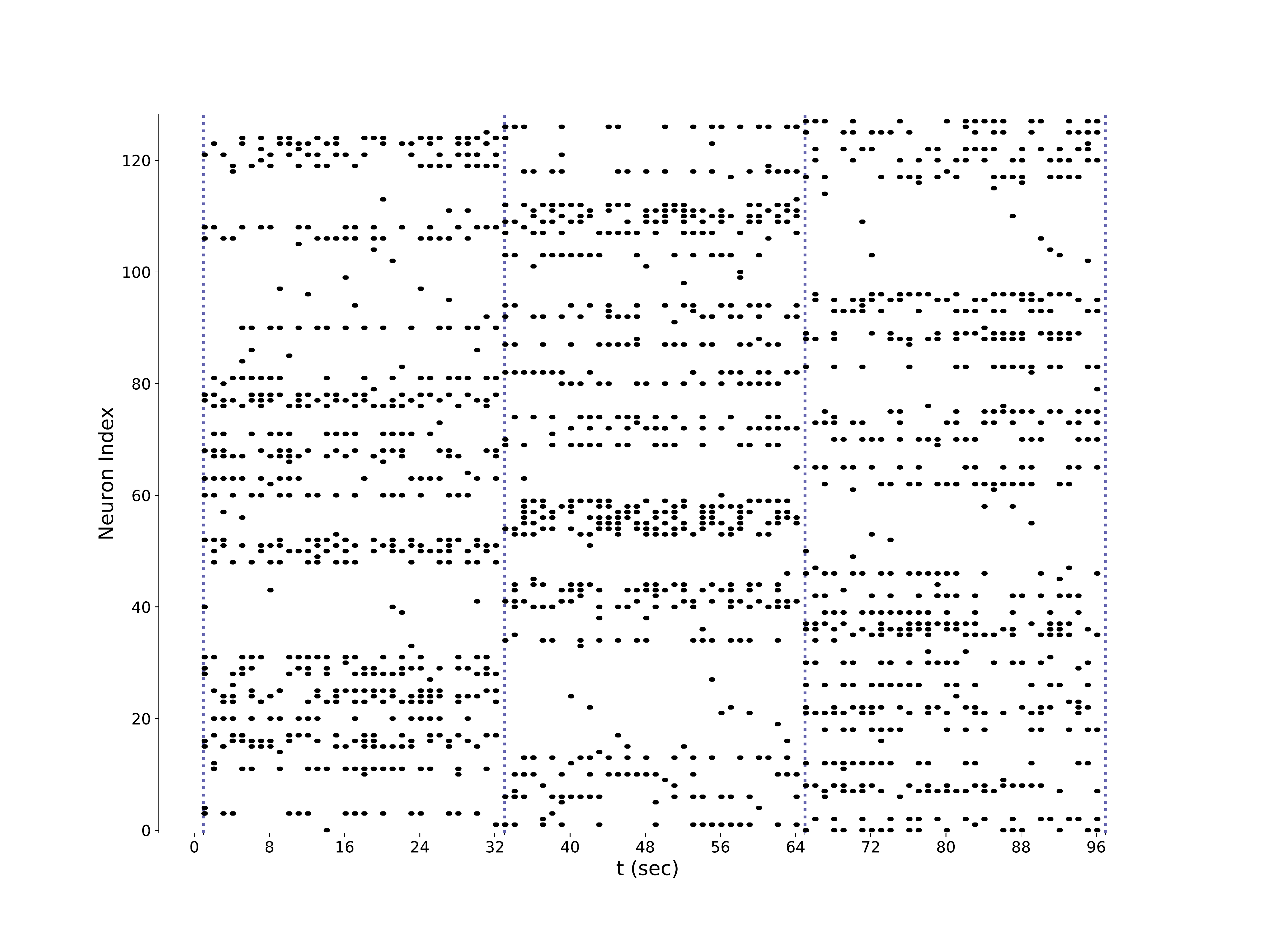}
\caption{The full spike raster for a Girvan-Newman benchmark graph instance. Only communities $Q_1$, $Q_2$ and $Q_4$ were driven. Each neuron in the communities was actively driven once. The vertical lines mark: $t= 1\; \mathrm{(sec)}$, the start of $Q_1$ driving; $t = 33 \; \mathrm{(sec)}$, the end of $Q_1$ driving and the start of $Q_2$ driving;  $t = 65 \; \mathrm{(sec)}$, the end of $Q_2$ driving and the start of $Q_4$ driving; and $t = 97 \; \mathrm{(sec)}$ the end of $Q_4$ driving.}
\label{fig:spike_raster}
\end{figure}
\begin{figure}
\includegraphics[width=1.0\columnwidth]{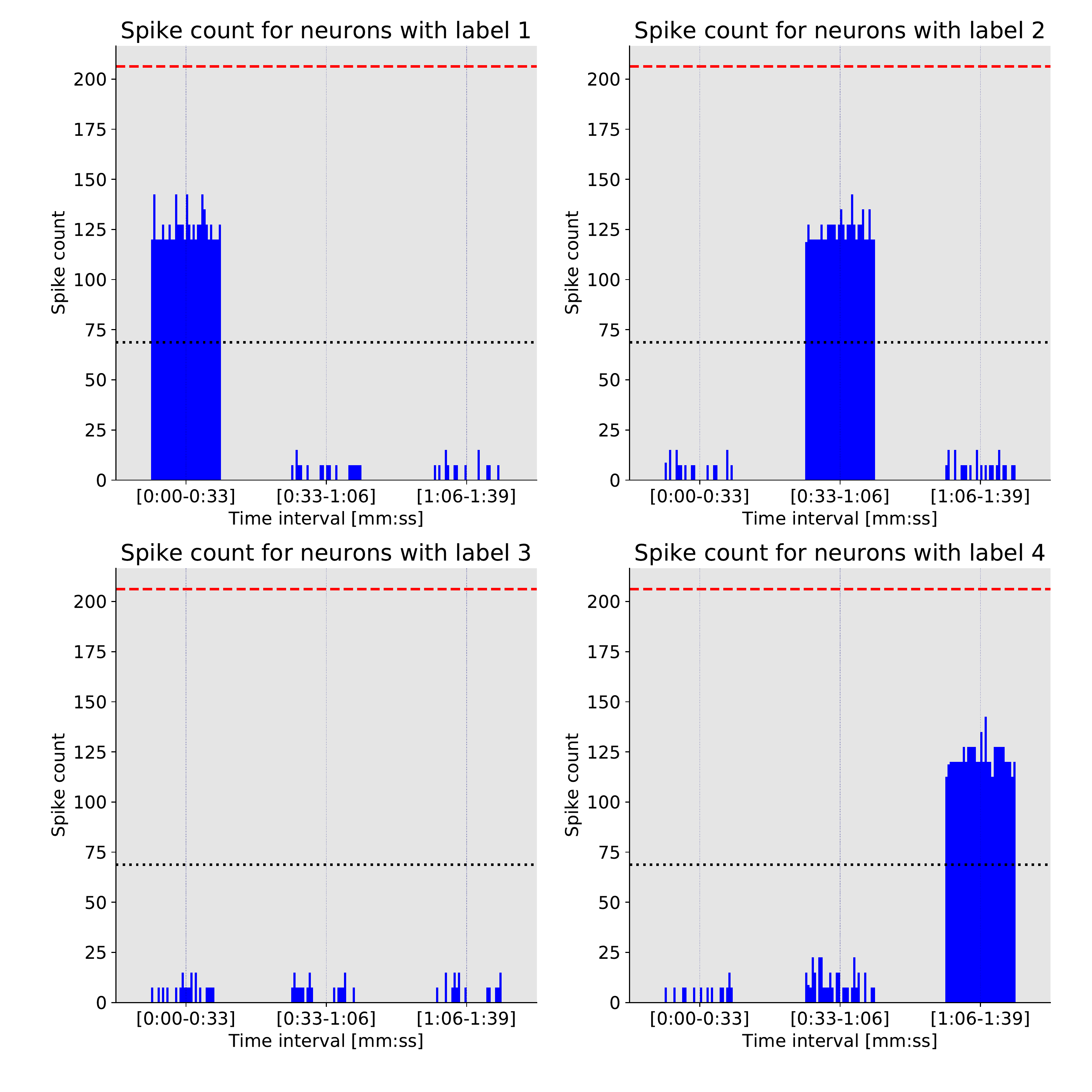}
\caption{The spike counts for neurons in each labelled community during three time windows of width $\Delta t \geq 32 \:\mathrm{seconds}$ showing how neurons in each community react when active driving is applied to neurons in $Q_1$, $Q_2$, $Q_4$ (respectively). The (red, dashed) horizontal is an upper bound on the total number of spikes that would be generated under square pulse driving of a $K_{32}$ community. The (black,dashed) line is a lower bound generated by using the mean degree $\langle d \rangle = 16$ and the assumption that at least half of a neuron's neighbors have the same community label.}
\label{fig:histogram}
\end{figure}
Inhibitory synapses impede the spread of spike cascades throughout the entire neuron system, as shown in Fig. \ref{fig:comparison_matrix} and \ref{fig:histogram}. For an instance of the Girvan-Newman benchmark graph it is shown that when $3$ of the $4$ communities are driven, the similarity between spike trains in the same driven community is significantly higher than the similarity between spike trains in different communities. Additionally, the un-driven community never exhibits a significant degree of spike response, showing that the spread of spiking synchronicity throughout the entire network is impeded. 

\section{Identifying unknown communities}
\label{sec:future_work}
Future development of spike-based community detection is dependent on how well spiking neuron systems can be incorporated into existing algorithms. There exists a near linear algorithm for community identification called ``label propagation,'' introduced in 2007 by Raghavan, Albert and Kumara \cite{raghavan2007near}. The similarity between this method and a Potts spin model have been noted \cite{tibely2008equivalence}, and we consider label propagation to be an algorithm that can incorporate spike-based data. However there are several obstacles to overcome before spiking neuron systems can be incorporated into real-world community identification tasks. 

First, we need to know how a neuron's spike response is affected when the neurons are driven in a random order. We use the Girvan-Newman benchmark graph shown in Fig. \ref{fig:girvan_newman_graph}, but instead of the community-ordered driving used in Sections \ref{sec:binary_code} and \ref{sec:bipolar_code}, we randomly permute and drive the entire neuron set. Randomly permuting the neuron set reduces the usefulness of bipolar decoding and the spiking data analysis in this section only uses binary decoding. Additionally, when randomly driving neurons, $\Delta t$ must be carefully chosen such that the response between subsequent driven neurons are not covered by a single time window. If that happens, then any distinction between signals belonging to individual neurons which may have different community labels may be lost. There is now an upper bound on the time window: $\Delta t < t_A$. 

Second, the scalability of the Hamming metric of Eq. \ref{eq:Hamming_metric_modified} is quite poor. Analysis of larger graphs will require driving a large number of neurons and the length of the binary decoded spike trains will rapidly increase.   As the length a binary decoded spike train increases, the similarity between two neurons that fire frequently will become quite large. To remedy this, we return to the original implementation of Hamming metric as defined in \cite{humphries2011spike}. 

\begin{figure}
\includegraphics[width=1.0\columnwidth]{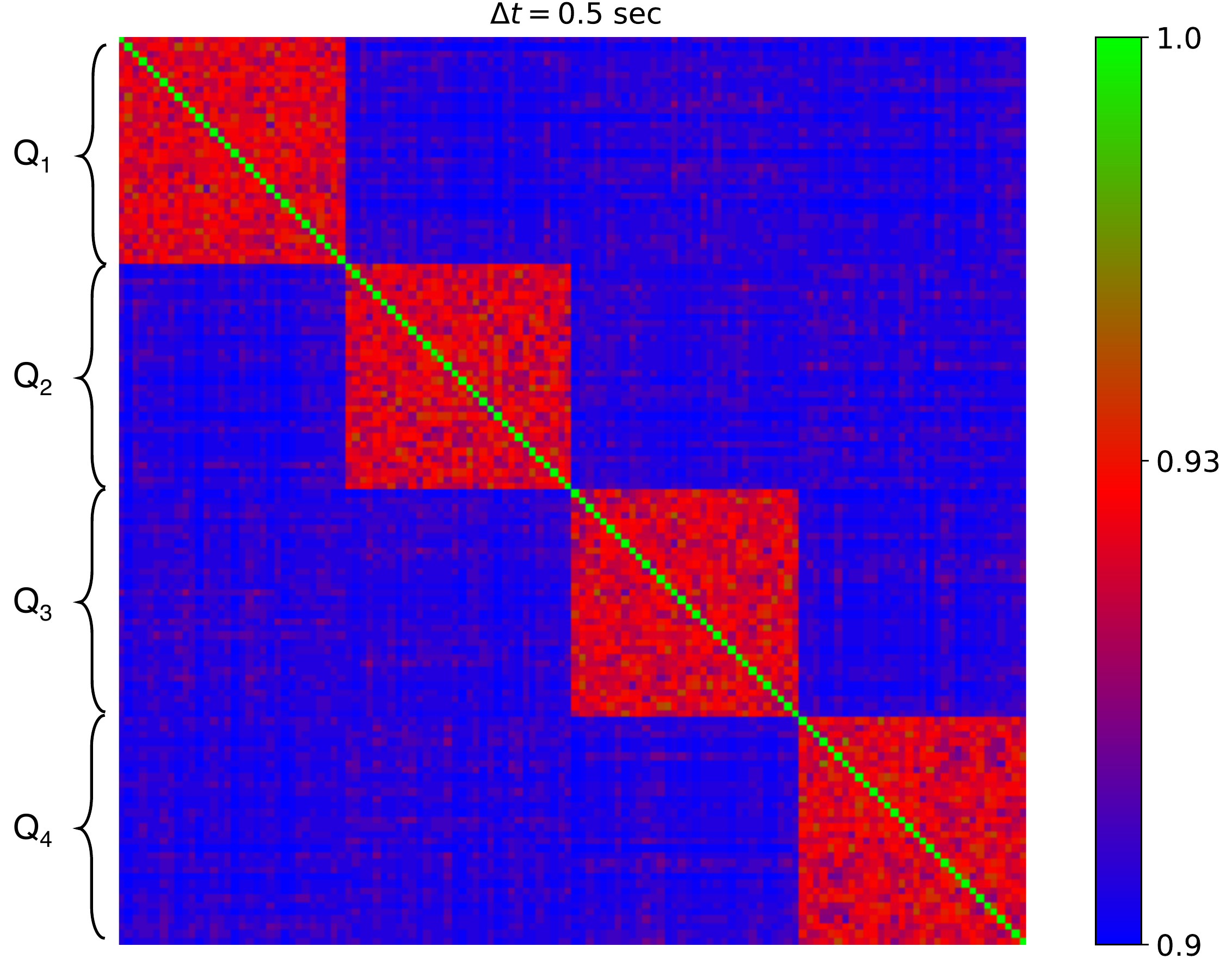}
\caption{The comparison metric of a Girvan-Newman graph instance on $128$ vertices: the complete neuron set is first randomly permuted, then every neuron is individually driven by a square pulse. The time window $\Delta t = 30 \:\mathrm{ms}$ was chosen such that $ \delta_1 < \Delta t < \delta_2 \ll t_A = 200 \:\mathrm{ms}$. The Hamming metric does not include any down-weighting terms, and the scale is now limited to the range $[0.0,1.0]$. Along the matrix diagonal, it is seen that $H(x_i,x_i)=1.00$.}
\label{fig:Comparison_matrix_2}
\end{figure}

\begin{figure*}
\centering
\includegraphics[width=1.0\textwidth]{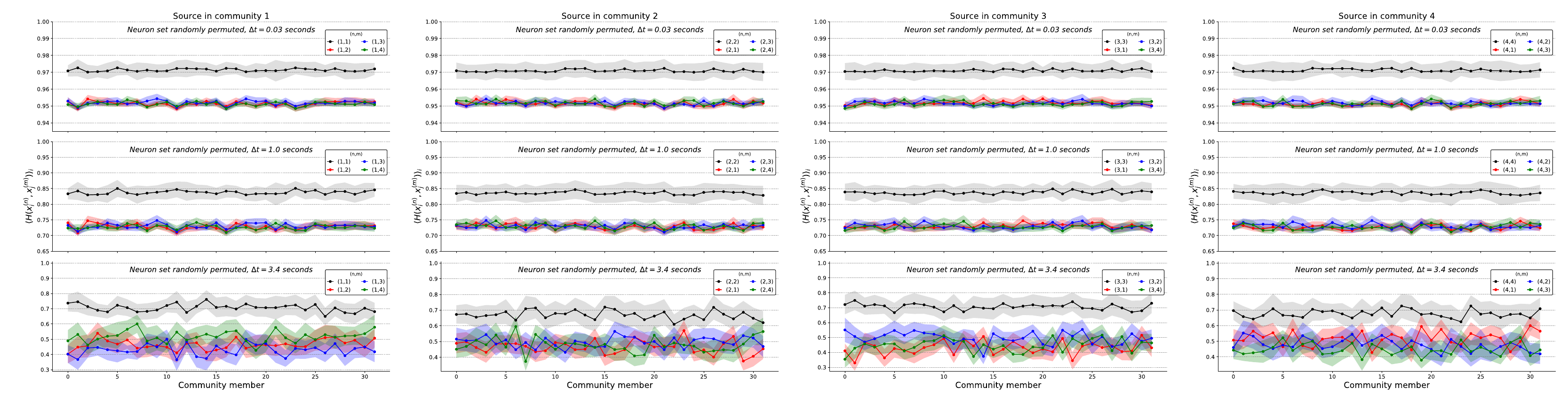}
\caption{For an instance of the Girvan-Newman graph on $128$ vertices, the complete neuron set is randomly permuted, then every neuron is individually driven by a square pulse. The mean Hamming metric $\langle H(x_{i}^{m},x_{j}^{m}) \rangle_i$ for source and target in the same community (black) can be distinguished from the mean Hamming metric mean Hamming metric $\langle H(x_{i}^{m},x_{j}^{n}) \rangle_i$ for source and target in different communities (red, blue, green) if the bin width $\Delta t $ is small. The standard deviation in the mean is shown by the shaded regions.}
\label{fig:Hamming_metric_2x2}
\end{figure*}

Returning to the Hamming metric as originally defined in \cite{humphries2011spike} does not affect the comparison matrix structure (see Fig. \ref{fig:Comparison_matrix_2}) and using the Hamming metric, we show how spiking data can be used in label propagation algorithms. Label propagation is done by choosing a vertex to act as a ``source,'' fixing its label, then updating the labels of all its neighboring vertices. For spiking data to be incorporated into this method, we need to ensure that the choice of source does not significantly impact the linear separability of the Hamming metric values. The role of $\Delta t$ in tuning this quality is shown in Fig. \ref{fig:Hamming_metric_2x2}: at $\Delta t = 3.2 \; \mathrm{(sec)}$ the mean Hamming metric value is defined as:
\begin{equation}
\langle H(x_{i}^{m},x_{j}^{n}) \rangle_i = \frac{1}{|Q_m|}\sum_{\substack{i \in Q_n \\ i \neq j}}H(x_{i}^{m},x_{j}^{n}).
\end{equation}
If $m=n$, the self-similarity value $H(x^{m}_i,x^{m}_i)=1.00$ is excluded. When the source and target are in different labelled communities ($n \neq m$) the mean metric is nearly overlapping, but when the source and target are in the same community, the mean is nearly $1.00$.  The linear separability of the mean metric value is dependent on the size of $\Delta t$, as seen in Fig. \ref{fig:Hamming_metric_2x2}.

\section{Conclusions}
Our approach to a community detection using spike-based computing depended on vertices within the same community exhibiting similar spiking patterns. We have studied how SNNs can be constructed, driven by external current, and decoded in order to generate such spiking patterns. An undirected graph is mapped to a fully connected set of homogeneous spiking neurons, with both excitatory and inhibitory synapses and selectively driven by square pulse currents. The neurons contained within densely connected regions exhibit synchronicity in their firing patterns.

Spike trains are decoded by time window binning, using varying window widths ($\Delta t$). For $\Delta t < t_{A}$ the resulting binned vectors are converted into binary codes, with entries $\lbrace 0,1 \rbrace$ only. From these binary vectors and a Hamming-distance based metric, we construct a comparison matrix which had dimension $n \times n$ for a graph of order $n$. The degree of similarity between spike trains can be quantified and used to infer the membership of communities on the original graph. For $\Delta t \approx |Q_i| (t_A + \delta_{pulse})$ the resulting vectors tabulate the number of spikes fired during $\Delta t$ and use a pre-determined threshold to determine the appropriate label to assign to the neurons.  

Using binary decoding, the driven communities could be distinguished by the Hamming metric similarity.  Simulations of spiking neuron dynamics demonstrated the ability to identify non-overlapping, communities of equal order (see Figs. \ref{fig:comparison_matrix}, \ref{fig:Comparison_matrix_2}) for graphs with known community structure. In Fig. \ref{fig:comparison_matrix}, it was seen that the use of inhibitory synaptic connections prevented any large scale spike cascade.  The Hamming metric can distinguish a single community when a neuron system is driven sequentially according to community, or when the neuron set is randomly permuted. Bipolar decoding, assigning community labels according to the spike counts, can distinguish the driven communities by the dominant spiking behavior. However, the bipolar decoding is only used when a neuron system is driven sequentially according to community. The results presented here are generated from one instance of the Girvan-Newman benchmark graph (shown in Fig. \ref{fig:girvan_newman_graph}), multiple instances have been tested with similar results. Additional simulations have also been tested on graphs with unequal-sized communities. We explore how our method performs on un-ordered data, by randomly permuting and driving the entire neuron set. In Fig. \ref{fig:Hamming_metric_2x2}, the linear separability of the Hamming metric value shows that for small enough $\Delta t$ value, the Hamming metric can distinguish one community from the remaining $3$ communities.

Our results show that spiking data can be used to identify communities in undirected graphs. The separability of the Hamming similarity measure (see Fig. \ref{fig:Hamming_metric_2x2}) could be incorporated into a label propagation workflow \cite{raghavan2007near,gallagher2008using,Peel2016relational}. Community detection is a unique use of neuromorphic hardware, utilizing neural systems that do not have hidden units, or require training over large data sets. 

Future work related to spike-based community detection will investigate two areas: the applicability, and the scalability of this method. Applicability focuses on quantifying the limits of a spike-based approach: effectiveness on other graph classes, identification of overlapping communities, and establishing a resolution limit (smallest community that can be identified in a large network). Scalability addresses how this method can be applied to large network analysis. In particular, as presented in this work, the binary analysis has the potential for poor scaling: a system of $n$ vertices is mapped to a densely connected system of $n$ neurons and produces a matrix of dimension $n \times n$.

\bibliographystyle{ACM-Reference-Format}
\bibliography{neuromorph_2017} 


\begin{thebibliography}{00}


\ifx \showCODEN    \undefined \def \showCODEN     #1{\unskip}     \fi
\ifx \showDOI      \undefined \def \showDOI       #1{#1}\fi
\ifx \showISBNx    \undefined \def \showISBNx     #1{\unskip}     \fi
\ifx \showISBNxiii \undefined \def \showISBNxiii  #1{\unskip}     \fi
\ifx \showISSN     \undefined \def \showISSN      #1{\unskip}     \fi
\ifx \showLCCN     \undefined \def \showLCCN      #1{\unskip}     \fi
\ifx \shownote     \undefined \def \shownote      #1{#1}          \fi
\ifx \showarticletitle \undefined \def \showarticletitle #1{#1}   \fi
\ifx \showURL      \undefined \def \showURL       {\relax}        \fi
\providecommand\bibfield[2]{#2}
\providecommand\bibinfo[2]{#2}
\providecommand\natexlab[1]{#1}
\providecommand\showeprint[2][]{arXiv:#2}

\bibitem[\protect\citeauthoryear{Blatt, Wiseman, and Domany}{Blatt
  et~al\mbox{.}}{1996}]%
        {PhysRevLett.76.3251}
\bibfield{author}{\bibinfo{person}{Marcelo Blatt}, \bibinfo{person}{Shai
  Wiseman}, {and} \bibinfo{person}{Eytan Domany}.}
  \bibinfo{year}{1996}\natexlab{}.
\newblock \showarticletitle{Superparamagnetic Clustering of Data}.
\newblock \bibinfo{journal}{{\em Phys. Rev. Lett.\/}}  \bibinfo{volume}{76}
  (\bibinfo{date}{Apr} \bibinfo{year}{1996}), \bibinfo{pages}{3251--3254}.
\newblock
Issue 18.
\showDOI{%
\url{https://doi.org/10.1103/PhysRevLett.76.3251}}


\bibitem[\protect\citeauthoryear{Boccaletti, Latora, Moreno, Chavez, and
  Hwang}{Boccaletti et~al\mbox{.}}{2006}]%
        {boccaletti2006complex}
\bibfield{author}{\bibinfo{person}{Stefano Boccaletti}, \bibinfo{person}{Vito
  Latora}, \bibinfo{person}{Yamir Moreno}, \bibinfo{person}{Martin Chavez},
  {and} \bibinfo{person}{D-U Hwang}.} \bibinfo{year}{2006}\natexlab{}.
\newblock \showarticletitle{Complex networks: Structure and dynamics}.
\newblock \bibinfo{journal}{{\em Physics reports\/}} \bibinfo{volume}{424},
  \bibinfo{number}{4} (\bibinfo{year}{2006}), \bibinfo{pages}{175--308}.
\newblock


\bibitem[\protect\citeauthoryear{Cassidy, Merolla, Arthur, Esser, Jackson,
  Alvarez-Icaza, Datta, Sawada, Wong, Feldman, et~al\mbox{.}}{Cassidy
  et~al\mbox{.}}{2013}]%
        {cassidy2013cognitive}
\bibfield{author}{\bibinfo{person}{Andrew~S Cassidy}, \bibinfo{person}{Paul
  Merolla}, \bibinfo{person}{John~V Arthur}, \bibinfo{person}{Steve~K Esser},
  \bibinfo{person}{Bryan Jackson}, \bibinfo{person}{Rodrigo Alvarez-Icaza},
  \bibinfo{person}{Pallab Datta}, \bibinfo{person}{Jun Sawada},
  \bibinfo{person}{Theodore~M Wong}, \bibinfo{person}{Vitaly Feldman},
  {et~al\mbox{.}}} \bibinfo{year}{2013}\natexlab{}.
\newblock \showarticletitle{Cognitive computing building block: A versatile and
  efficient digital neuron model for neurosynaptic cores}. In
  \bibinfo{booktitle}{{\em Neural Networks (IJCNN), The 2013 International
  Joint Conference on}}. IEEE, \bibinfo{pages}{1--10}.
\newblock


\bibitem[\protect\citeauthoryear{Clauset, Newman, and Moore}{Clauset
  et~al\mbox{.}}{2004}]%
        {PhysRevE.70.066111}
\bibfield{author}{\bibinfo{person}{Aaron Clauset}, \bibinfo{person}{M.~E.~J.
  Newman}, {and} \bibinfo{person}{Cristopher Moore}.}
  \bibinfo{year}{2004}\natexlab{}.
\newblock \showarticletitle{Finding community structure in very large
  networks}.
\newblock \bibinfo{journal}{{\em Phys. Rev. E\/}}  \bibinfo{volume}{70}
  (\bibinfo{date}{Dec} \bibinfo{year}{2004}), \bibinfo{pages}{066111}.
\newblock
Issue 6.
\showDOI{%
\url{https://doi.org/10.1103/PhysRevE.70.066111}}


\bibitem[\protect\citeauthoryear{Fortunato}{Fortunato}{2010}]%
        {fortunato2010community}
\bibfield{author}{\bibinfo{person}{Santo Fortunato}.}
  \bibinfo{year}{2010}\natexlab{}.
\newblock \showarticletitle{Community detection in graphs}.
\newblock \bibinfo{journal}{{\em Physics reports\/}} \bibinfo{volume}{486},
  \bibinfo{number}{3} (\bibinfo{year}{2010}), \bibinfo{pages}{75--174}.
\newblock


\bibitem[\protect\citeauthoryear{Fortunato}{Fortunato}{2017}]%
        {fortunato_site}
\bibfield{author}{\bibinfo{person}{Santo Fortunato}.}
  \bibinfo{year}{2017}\natexlab{}.
\newblock \bibinfo{title}{Santo Fortunato's Website: Software}.
\newblock
  \bibinfo{howpublished}{\url{https://sites.google.com/site/santofortunato/inthepress2}}.
    (\bibinfo{year}{2017}).
\newblock
\newblock
\shownote{Accessed: 2017-05-10.}


\bibitem[\protect\citeauthoryear{Gallagher, Tong, Eliassi-Rad, and
  Faloutsos}{Gallagher et~al\mbox{.}}{2008}]%
        {gallagher2008using}
\bibfield{author}{\bibinfo{person}{Brian Gallagher}, \bibinfo{person}{Hanghang
  Tong}, \bibinfo{person}{Tina Eliassi-Rad}, {and} \bibinfo{person}{Christos
  Faloutsos}.} \bibinfo{year}{2008}\natexlab{}.
\newblock \showarticletitle{Using Ghost Edges for Classification in Sparsely
  Labeled Networks}. In \bibinfo{booktitle}{{\em Proceedings of the 14th ACM
  SIGKDD International Conference on Knowledge Discovery and Data Mining}} {\em
  (\bibinfo{series}{KDD '08})}. \bibinfo{publisher}{ACM}, \bibinfo{address}{New
  York, NY, USA}, \bibinfo{pages}{256--264}.
\newblock
\showISBNx{978-1-60558-193-4}
\showDOI{%
\url{https://doi.org/10.1145/1401890.1401925}}


\bibitem[\protect\citeauthoryear{Gerstner and Kistler}{Gerstner and
  Kistler}{2002}]%
        {gerstner2002spiking}
\bibfield{author}{\bibinfo{person}{Wulfram Gerstner} {and}
  \bibinfo{person}{Werner~M Kistler}.} \bibinfo{year}{2002}\natexlab{}.
\newblock \bibinfo{booktitle}{{\em Spiking neuron models: Single neurons,
  populations, plasticity}}.
\newblock \bibinfo{publisher}{Cambridge university press}.
\newblock


\bibitem[\protect\citeauthoryear{Girvan and Newman}{Girvan and Newman}{2002}]%
        {girvan2002community}
\bibfield{author}{\bibinfo{person}{Michelle Girvan} {and}
  \bibinfo{person}{Mark~EJ Newman}.} \bibinfo{year}{2002}\natexlab{}.
\newblock \showarticletitle{Community structure in social and biological
  networks}.
\newblock \bibinfo{journal}{{\em Proceedings of the national academy of
  sciences\/}} \bibinfo{volume}{99}, \bibinfo{number}{12}
  (\bibinfo{year}{2002}), \bibinfo{pages}{7821--7826}.
\newblock


\bibitem[\protect\citeauthoryear{Goodman and Brette}{Goodman and
  Brette}{2008}]%
        {goodman2008brian}
\bibfield{author}{\bibinfo{person}{Dan Goodman} {and} \bibinfo{person}{Romain
  Brette}.} \bibinfo{year}{2008}\natexlab{}.
\newblock \showarticletitle{Brian: A Simulator for Spiking Neural Networks in
  Python}.
\newblock \bibinfo{journal}{{\em Frontiers in Neuroinformatics\/}}
  \bibinfo{volume}{2} (\bibinfo{year}{2008}).
\newblock


\bibitem[\protect\citeauthoryear{Hamilton, Imam, and Humble}{Hamilton
  et~al\mbox{.}}{2017}]%
        {Hamilton2017ns17}
\bibfield{author}{\bibinfo{person}{Kathleen~E. Hamilton},
  \bibinfo{person}{Neena Imam}, {and} \bibinfo{person}{Travis~S. Humble}.}
  \bibinfo{year}{2017}\natexlab{}.
\newblock \bibinfo{title}{Community Identification with spiking neural
  networks}.
\newblock \bibinfo{howpublished}{Poster presented at SIAM's Network Science
  Workshop 2017 July 13--14, Pittsburgh, PA.}.   (\bibinfo{year}{2017}).
\newblock


\bibitem[\protect\citeauthoryear{Hertz, Krogh, and Palmer}{Hertz
  et~al\mbox{.}}{1991}]%
        {hertz1991introduction}
\bibfield{author}{\bibinfo{person}{John Hertz}, \bibinfo{person}{Anders Krogh},
  {and} \bibinfo{person}{Richard~G Palmer}.} \bibinfo{year}{1991}\natexlab{}.
\newblock \bibinfo{booktitle}{{\em Introduction to the theory of neural
  computation}}. \bibinfo{series}{Santa Fe Institute studies in the sciences of
  complexity}, Vol.~\bibinfo{volume}{1}.
\newblock \bibinfo{publisher}{Addison-Wesley}.
\newblock


\bibitem[\protect\citeauthoryear{Hopfield}{Hopfield}{1982}]%
        {hopfield1982neural}
\bibfield{author}{\bibinfo{person}{John~J Hopfield}.}
  \bibinfo{year}{1982}\natexlab{}.
\newblock \showarticletitle{Neural networks and physical systems with emergent
  collective computational abilities}.
\newblock \bibinfo{journal}{{\em Proceedings of the national academy of
  sciences\/}} \bibinfo{volume}{79}, \bibinfo{number}{8}
  (\bibinfo{year}{1982}), \bibinfo{pages}{2554--2558}.
\newblock


\bibitem[\protect\citeauthoryear{Hopfield}{Hopfield}{1984}]%
        {hopfield1984neurons}
\bibfield{author}{\bibinfo{person}{John~J Hopfield}.}
  \bibinfo{year}{1984}\natexlab{}.
\newblock \showarticletitle{Neurons with graded response have collective
  computational properties like those of two-state neurons}.
\newblock \bibinfo{journal}{{\em Proceedings of the national academy of
  sciences\/}} \bibinfo{volume}{81}, \bibinfo{number}{10}
  (\bibinfo{year}{1984}), \bibinfo{pages}{3088--3092}.
\newblock


\bibitem[\protect\citeauthoryear{Hopfield and Tank}{Hopfield and Tank}{1985}]%
        {hopfield1985neural}
\bibfield{author}{\bibinfo{person}{John~J Hopfield} {and}
  \bibinfo{person}{David~W Tank}.} \bibinfo{year}{1985}\natexlab{}.
\newblock \showarticletitle{Neural computation of decisions in optimization
  problems}.
\newblock \bibinfo{journal}{{\em Biological cybernetics\/}}
  \bibinfo{volume}{52}, \bibinfo{number}{3} (\bibinfo{year}{1985}),
  \bibinfo{pages}{141--152}.
\newblock


\bibitem[\protect\citeauthoryear{Humphries}{Humphries}{2011}]%
        {humphries2011spike}
\bibfield{author}{\bibinfo{person}{Mark~D Humphries}.}
  \bibinfo{year}{2011}\natexlab{}.
\newblock \showarticletitle{Spike-train communities: finding groups of similar
  spike trains}.
\newblock \bibinfo{journal}{{\em Journal of Neuroscience\/}}
  \bibinfo{volume}{31}, \bibinfo{number}{6} (\bibinfo{year}{2011}),
  \bibinfo{pages}{2321--2336}.
\newblock


\bibitem[\protect\citeauthoryear{Kanamaru and Okabe}{Kanamaru and
  Okabe}{2000}]%
        {kanamaru2000associative}
\bibfield{author}{\bibinfo{person}{Takashi Kanamaru} {and}
  \bibinfo{person}{Yoichi Okabe}.} \bibinfo{year}{2000}\natexlab{}.
\newblock \showarticletitle{Associative memory retrieval induced by
  fluctuations in a pulsed neural network}.
\newblock \bibinfo{journal}{{\em Physical Review E\/}} \bibinfo{volume}{62},
  \bibinfo{number}{2} (\bibinfo{year}{2000}), \bibinfo{pages}{2629}.
\newblock


\bibitem[\protect\citeauthoryear{Kanamaru and Okabe}{Kanamaru and
  Okabe}{2001}]%
        {kanamaru2001fluctuation}
\bibfield{author}{\bibinfo{person}{Takashi Kanamaru} {and}
  \bibinfo{person}{Yoichi Okabe}.} \bibinfo{year}{2001}\natexlab{}.
\newblock \showarticletitle{Fluctuation-induced memory retrieval in a pulsed
  neural network storing sparse patterns with hierarchical correlations}.
\newblock \bibinfo{journal}{{\em Physical Review E\/}} \bibinfo{volume}{64},
  \bibinfo{number}{3} (\bibinfo{year}{2001}), \bibinfo{pages}{031904}.
\newblock


\bibitem[\protect\citeauthoryear{Lancichinetti and Fortunato}{Lancichinetti and
  Fortunato}{2009}]%
        {lancichinetti2009benchmarks}
\bibfield{author}{\bibinfo{person}{Andrea Lancichinetti} {and}
  \bibinfo{person}{Santo Fortunato}.} \bibinfo{year}{2009}\natexlab{}.
\newblock \showarticletitle{Benchmarks for testing community detection
  algorithms on directed and weighted graphs with overlapping communities}.
\newblock \bibinfo{journal}{{\em Physical Review E\/}} \bibinfo{volume}{80},
  \bibinfo{number}{1} (\bibinfo{year}{2009}), \bibinfo{pages}{016118}.
\newblock


\bibitem[\protect\citeauthoryear{Lancichinetti, Fortunato, and
  Radicchi}{Lancichinetti et~al\mbox{.}}{2008}]%
        {lancichinetti2008benchmark}
\bibfield{author}{\bibinfo{person}{Andrea Lancichinetti},
  \bibinfo{person}{Santo Fortunato}, {and} \bibinfo{person}{Filippo Radicchi}.}
  \bibinfo{year}{2008}\natexlab{}.
\newblock \showarticletitle{Benchmark graphs for testing community detection
  algorithms}.
\newblock \bibinfo{journal}{{\em Physical review E\/}} \bibinfo{volume}{78},
  \bibinfo{number}{4} (\bibinfo{year}{2008}), \bibinfo{pages}{046110}.
\newblock


\bibitem[\protect\citeauthoryear{Lowel and Singer}{Lowel and Singer}{1992}]%
        {Lowel209}
\bibfield{author}{\bibinfo{person}{S Lowel} {and} \bibinfo{person}{W Singer}.}
  \bibinfo{year}{1992}\natexlab{}.
\newblock \showarticletitle{Selection of intrinsic horizontal connections in
  the visual cortex by correlated neuronal activity}.
\newblock \bibinfo{journal}{{\em Science\/}} \bibinfo{volume}{255},
  \bibinfo{number}{5041} (\bibinfo{year}{1992}), \bibinfo{pages}{209--212}.
\newblock
\showISSN{0036-8075}
\showDOI{%
\url{https://doi.org/10.1126/science.1372754}}
\showeprint{http://science.sciencemag.org/content/255/5041/209.full.pdf}


\bibitem[\protect\citeauthoryear{Maass and Bishop}{Maass and Bishop}{2001}]%
        {maass2001pulsed}
\bibfield{author}{\bibinfo{person}{Wolfgang Maass} {and}
  \bibinfo{person}{Christopher~M Bishop}.} \bibinfo{year}{2001}\natexlab{}.
\newblock \bibinfo{booktitle}{{\em Pulsed neural networks}}.
\newblock \bibinfo{publisher}{MIT press}.
\newblock


\bibitem[\protect\citeauthoryear{Maass and Natschl{\"a}ger}{Maass and
  Natschl{\"a}ger}{1997}]%
        {maass1997networks}
\bibfield{author}{\bibinfo{person}{Wolfgang Maass} {and}
  \bibinfo{person}{Thomas Natschl{\"a}ger}.} \bibinfo{year}{1997}\natexlab{}.
\newblock \showarticletitle{Networks of spiking neurons can emulate arbitrary
  {H}opfield nets in temporal coding}.
\newblock \bibinfo{journal}{{\em Network: Computation in Neural Systems\/}}
  \bibinfo{volume}{8}, \bibinfo{number}{4} (\bibinfo{year}{1997}),
  \bibinfo{pages}{355--371}.
\newblock


\bibitem[\protect\citeauthoryear{Malliaros and Vazirgiannis}{Malliaros and
  Vazirgiannis}{2013}]%
        {malliaros2013clustering}
\bibfield{author}{\bibinfo{person}{Fragkiskos~D Malliaros} {and}
  \bibinfo{person}{Michalis Vazirgiannis}.} \bibinfo{year}{2013}\natexlab{}.
\newblock \showarticletitle{Clustering and community detection in directed
  networks: A survey}.
\newblock \bibinfo{journal}{{\em Physics Reports\/}} \bibinfo{volume}{533},
  \bibinfo{number}{4} (\bibinfo{year}{2013}), \bibinfo{pages}{95--142}.
\newblock


\bibitem[\protect\citeauthoryear{Merolla, Arthur, Alvarez-Icaza, Cassidy,
  Sawada, Akopyan, Jackson, Imam, Guo, Nakamura, et~al\mbox{.}}{Merolla
  et~al\mbox{.}}{2014}]%
        {merolla2014million}
\bibfield{author}{\bibinfo{person}{Paul~A Merolla}, \bibinfo{person}{John~V
  Arthur}, \bibinfo{person}{Rodrigo Alvarez-Icaza}, \bibinfo{person}{Andrew~S
  Cassidy}, \bibinfo{person}{Jun Sawada}, \bibinfo{person}{Filipp Akopyan},
  \bibinfo{person}{Bryan~L Jackson}, \bibinfo{person}{Nabil Imam},
  \bibinfo{person}{Chen Guo}, \bibinfo{person}{Yutaka Nakamura},
  {et~al\mbox{.}}} \bibinfo{year}{2014}\natexlab{}.
\newblock \showarticletitle{A million spiking-neuron integrated circuit with a
  scalable communication network and interface}.
\newblock \bibinfo{journal}{{\em Science\/}} \bibinfo{volume}{345},
  \bibinfo{number}{6197} (\bibinfo{year}{2014}), \bibinfo{pages}{668--673}.
\newblock


\bibitem[\protect\citeauthoryear{Newman and Girvan}{Newman and Girvan}{2004}]%
        {newman2004finding}
\bibfield{author}{\bibinfo{person}{Mark~EJ Newman} {and}
  \bibinfo{person}{Michelle Girvan}.} \bibinfo{year}{2004}\natexlab{}.
\newblock \showarticletitle{Finding and evaluating community structure in
  networks}.
\newblock \bibinfo{journal}{{\em Physical review E\/}} \bibinfo{volume}{69},
  \bibinfo{number}{2} (\bibinfo{year}{2004}), \bibinfo{pages}{026113}.
\newblock


\bibitem[\protect\citeauthoryear{Newman}{Newman}{2010}]%
        {NewmanNetworks}
\bibfield{author}{\bibinfo{person}{M.~E.~J. Newman}.}
  \bibinfo{year}{2010}\natexlab{}.
\newblock \bibinfo{booktitle}{{\em Networks: An Introduction}}.
\newblock \bibinfo{publisher}{Oxford University Press}.
\newblock


\bibitem[\protect\citeauthoryear{{Peel}}{{Peel}}{2016}]%
        {Peel2016relational}
\bibfield{author}{\bibinfo{person}{L. {Peel}}.}
  \bibinfo{year}{2016}\natexlab{}.
\newblock \showarticletitle{{Graph-based semi-supervised learning for
  relational networks}}.
\newblock \bibinfo{journal}{{\em ArXiv e-prints\/}} (\bibinfo{date}{Dec.}
  \bibinfo{year}{2016}).
\newblock
\showeprint[arxiv]{1612.05001}


\bibitem[\protect\citeauthoryear{Peel, Larremore, and Clauset}{Peel
  et~al\mbox{.}}{2017}]%
        {peel2017ground}
\bibfield{author}{\bibinfo{person}{Leto Peel}, \bibinfo{person}{Daniel~B
  Larremore}, {and} \bibinfo{person}{Aaron Clauset}.}
  \bibinfo{year}{2017}\natexlab{}.
\newblock \showarticletitle{The ground truth about metadata and community
  detection in networks}.
\newblock \bibinfo{journal}{{\em Science Advances\/}} \bibinfo{volume}{3},
  \bibinfo{number}{5} (\bibinfo{year}{2017}), \bibinfo{pages}{e1602548}.
\newblock


\bibitem[\protect\citeauthoryear{Raghavan, Albert, and Kumara}{Raghavan
  et~al\mbox{.}}{2007}]%
        {raghavan2007near}
\bibfield{author}{\bibinfo{person}{Usha~Nandini Raghavan},
  \bibinfo{person}{R{\'e}ka Albert}, {and} \bibinfo{person}{Soundar Kumara}.}
  \bibinfo{year}{2007}\natexlab{}.
\newblock \showarticletitle{Near linear time algorithm to detect community
  structures in large-scale networks}.
\newblock \bibinfo{journal}{{\em Physical Review E\/}} \bibinfo{volume}{76},
  \bibinfo{number}{3} (\bibinfo{year}{2007}), \bibinfo{pages}{036106}.
\newblock


\bibitem[\protect\citeauthoryear{Reichardt and Bornholdt}{Reichardt and
  Bornholdt}{2004}]%
        {reichardt2004detecting}
\bibfield{author}{\bibinfo{person}{J{\"o}rg Reichardt} {and}
  \bibinfo{person}{Stefan Bornholdt}.} \bibinfo{year}{2004}\natexlab{}.
\newblock \showarticletitle{Detecting fuzzy community structures in complex
  networks with a {P}otts model}.
\newblock \bibinfo{journal}{{\em Physical Review Letters\/}}
  \bibinfo{volume}{93}, \bibinfo{number}{21} (\bibinfo{year}{2004}),
  \bibinfo{pages}{218701}.
\newblock


\bibitem[\protect\citeauthoryear{Reichardt and Bornholdt}{Reichardt and
  Bornholdt}{2006}]%
        {PhysRevE.74.016110}
\bibfield{author}{\bibinfo{person}{J\"org Reichardt} {and}
  \bibinfo{person}{Stefan Bornholdt}.} \bibinfo{year}{2006}\natexlab{}.
\newblock \showarticletitle{Statistical mechanics of community detection}.
\newblock \bibinfo{journal}{{\em Phys. Rev. E\/}}  \bibinfo{volume}{74}
  (\bibinfo{date}{Jul} \bibinfo{year}{2006}), \bibinfo{pages}{016110}.
\newblock
Issue 1.
\showDOI{%
\url{https://doi.org/10.1103/PhysRevE.74.016110}}


\bibitem[\protect\citeauthoryear{Schaub, Delvenne, Rosvall, and
  Lambiotte}{Schaub et~al\mbox{.}}{2017}]%
        {Schaub2017}
\bibfield{author}{\bibinfo{person}{Michael~T. Schaub},
  \bibinfo{person}{Jean-Charles Delvenne}, \bibinfo{person}{Martin Rosvall},
  {and} \bibinfo{person}{Renaud Lambiotte}.} \bibinfo{year}{2017}\natexlab{}.
\newblock \showarticletitle{The many facets of community detection in complex
  networks}.
\newblock \bibinfo{journal}{{\em Applied Network Science\/}}
  \bibinfo{volume}{2}, \bibinfo{number}{1} (\bibinfo{year}{2017}),
  \bibinfo{pages}{4}.
\newblock
\showISSN{2364-8228}
\showDOI{%
\url{https://doi.org/10.1007/s41109-017-0023-6}}


\bibitem[\protect\citeauthoryear{Tanaka, Morie, and Aihara}{Tanaka
  et~al\mbox{.}}{2005}]%
        {tanaka2005associative}
\bibfield{author}{\bibinfo{person}{Hideki Tanaka}, \bibinfo{person}{Takashi
  Morie}, {and} \bibinfo{person}{Kazuyuki Aihara}.}
  \bibinfo{year}{2005}\natexlab{}.
\newblock \showarticletitle{Associative memory operation in a {H}opfield-type
  spiking neural network with modulation of resting membrane potential}. In
  \bibinfo{booktitle}{{\em Int. Symp. on Nonlinear Theory and its Applications,
  Bruges, Belgium}}.
\newblock


\bibitem[\protect\citeauthoryear{Tib{\'e}ly and Kert{\'e}sz}{Tib{\'e}ly and
  Kert{\'e}sz}{2008}]%
        {tibely2008equivalence}
\bibfield{author}{\bibinfo{person}{Gergely Tib{\'e}ly} {and}
  \bibinfo{person}{J{\'a}nos Kert{\'e}sz}.} \bibinfo{year}{2008}\natexlab{}.
\newblock \showarticletitle{On the equivalence of the label propagation method
  of community detection and a {P}otts model approach}.
\newblock \bibinfo{journal}{{\em Physica A: Statistical Mechanics and its
  Applications\/}} \bibinfo{volume}{387}, \bibinfo{number}{19}
  (\bibinfo{year}{2008}), \bibinfo{pages}{4982--4984}.
\newblock


\bibitem[\protect\citeauthoryear{Van Den~Bout and Miller}{Van Den~Bout and
  Miller}{1990}]%
        {van1990graph}
\bibfield{author}{\bibinfo{person}{David~E Van Den~Bout} {and}
  \bibinfo{person}{Thomas~K Miller}.} \bibinfo{year}{1990}\natexlab{}.
\newblock \showarticletitle{Graph partitioning using annealed neural networks}.
\newblock \bibinfo{journal}{{\em IEEE Transactions on neural networks\/}}
  \bibinfo{volume}{1}, \bibinfo{number}{2} (\bibinfo{year}{1990}),
  \bibinfo{pages}{192--203}.
\newblock


\end{thebibliography}

\end{document}